\documentclass[letterpaper, 10 pt, conference]{ieeeconf}  
\usepackage{caption}

\IEEEoverridecommandlockouts                              

\usepackage{graphicx} 
\usepackage{amsfonts}
\usepackage{amsmath,bm}
\usepackage{caption}

\usepackage{amssymb}
\usepackage{subcaption}
\usepackage{enumerate}

\graphicspath{ {./images/} }
\usepackage[english]{babel}

\author{Share\LaTeX}
\title{\LARGE \bf Non-Normalized Solutions of Generalized Nash Equilibrium in Autonomous Racing}

\author{Mark Pustilnik$^{1}$, Antonio Loquercio$^{2}$, Francesco Borrelli$^{1}$
\thanks{$^{1}$University of California at Berkeley, USA \quad {\tt\small \{pkmark, fborrelli\}@berkeley.edu}}%
\thanks{$^{2}$University of Pennsylvania, USA \quad {\tt\small aloque@seas.upenn.edu}}%
}

\date{March 2025}

\begin{document}

\maketitle
\thispagestyle{empty}
\pagestyle{empty}

\begin{abstract}

In dynamic games with shared constraints, Generalized Nash Equilibria (GNE) are often computed using the normalized solution concept, which assumes identical Lagrange multipliers for shared constraints across all players. While widely used, this approach excludes other potentially valuable GNE.  
This paper addresses the limitations of normalized solutions in racing scenarios through three key contributions. First, we highlight the shortcomings of normalized solutions with a simple racing example. Second, we propose a novel method based on the Mixed Complementarity Problem (MCP) formulation to compute non-normalized Generalized Nash Equilibria (GNE). Third, we demonstrate that our proposed method overcomes the limitations of normalized GNE solutions and enables richer multi-modal interactions in realistic racing scenarios. 
\end{abstract}


\section{Introduction}
Computing a Nash Equilibrium (NE) in a Dynamic Game (DG) is a well-established concept. In DGs, players often share constraints, making each player's feasible strategy set dependent on others’ strategies. This extends the problem to a Generalized Nash Equilibrium Problem (GNEP). NE serves as a fundamental tool for analyzing competitive behavior across various domains, including economics \cite{arrow2024existence} and racing \cite{zhu2023sequential, spica2020real}.

In general, the existence or uniqueness of a GNE is not guaranteed.
We are interested in problems where GNEPs admit multiple solutions. The standard approach for solving GNEPs is to derive and combine the Karush-Kuhn-Tucker (KKT) conditions for each player’s optimization problem~\cite{dreves2011solution}. When shared constraints exist, the Lagrange multipliers associated with these constraints are typically assumed to be identical for all players. This concept, known as the ``normalized solution", was introduced by Rosen \cite{rosen1965existence}, who demonstrated its existence and uniqueness under specific conditions. This solution also arises from reformulating the DG as a Quasi Variational Inequality (QVI) problem. Harker \cite{harker1991generalized} and others \cite{facchinei2007generalized} further showed that this reformulation reduces the solution set to the normalized solution, provided it exists. The normalized solution is widely adopted in various applications, including racing \cite{zhu2023sequential, liu2023learning, spica2020real}, economics \cite{bacsar1998dynamic}, and traffic systems \cite{migot2019revisiting}, and has become the standard approach for solving DGs due to its computational, numerical, and conceptual advantages. Despite these benefits, the normalized solution may not always capture the complexity of real-world interactions, where other GNEs could better represent asymmetric or more intuitive behavior. However, computing non-normalized GNEs is challenging. Standard approaches to solving GNEPs, such as reformulating the KKT conditions as a QVI is inherently bias toward normalized solutions \cite{facchinei2007generalized}. Reformulating the problem as Mixed Complementarity Problem (MCP) is practically also biased toward the normalized solution. In fact many numerical solvers converge to normalized solutions even when no such constraint is imposed on the solution space. This paper tackles these limitations through three key contributions:
1) highlights limits of normalized solutions in racing scenarios with a simple example, 2) Proposing a method that extends the MCP framework to calculate non-normalized GNEs and 3) show that the propose method overcome such limitations and generates interesting multi modal interactions in realistic racing scenarios.


\section{Problem Formulation} \label{Problem}
Given $M$ players, each player $i \in \{1,...,M\}$ controls the variables $x^i \in \mathbb{R}^{n_i}$. 
The vector $x \in \mathbb{R}^n$ is formed by concatenating all the players' decision variables:

\begin{equation} \label{x_vec}
    x := [(x^1)^T, ..., (x^M)^T]^T
\end{equation}
where $n := \sum^M_{i=1}n_i$. $x^{-i}$ represent the decision variables of all players except player $i$. To emphasize the decision variables of player $i$ within $x$ we write $(x^i,x^{-i})$ instead of $x$. The aim of player $i$ is to choose the $x^i$ which minimizes its own cost function $J_i(x^i,x^{-i})$ refer to such decision as \emph{strategy}.
The feasible set of the $i$-th decision depends on the strategies of the other players:
\begin{equation} \label{game_def}
    \min_{x^i} {J_i(x^i,x^{-i})} \quad \text{subject to} \quad x^i \in \mathcal{X}_i(x^{-i}) \subseteq \mathbb{R}^{n_i}
\end{equation}
The feasible set of each player can be represented in the following form:
\begin{align}
    \mathcal{X}_i(x_{-i})& := \{x_i \in \mathbb{R}^{n_i} | h_i(x_i) = 0, \nonumber\\
     &,g_i(x_i) \leq 0, s(x_i,x_{-i}) \leq 0\}
\end{align}
where, $h_i: \mathbb{R}^{n_i} \to \mathbb{R}^{k_i}$ defines \emph{private} equality constraints that depend only on player $i$'s strategy, $g_i:\mathbb{R}^{n_i} \rightarrow \mathbb{R}^{m_i}$ defines the \emph{private} inequality constraints of player $i$ that depends only on player $i$'s strategy. $s:\mathbb{R}^{n} \rightarrow \mathbb{R}^{m_0}$ defines the \emph{shared} constraints that depends on the strategies of all players and shared by all players. The solution set of problem (\ref{game_def}) for player $i$ is denoted by $\mathcal{S}_i(\bar{x}^{-i})$. The Generalized Nash Equilibrium Problem (GNEP) is the problem of finding vector $\bar{x}$ such that:
\begin{equation} \label{game_sol}
    \bar{x}^i \in \mathcal{S}_i(\bar{x}^{-i}) \quad \forall i \in \{1,...,M\}
\end{equation}
A solution to the GNEP is the GNE. A GNE is the set of strategies that characterized by the fact that none of the players can improve \emph{unilaterally} it's cost function by changing its strategy in a feasible direction. Furthermore, it is typical to assume the following assumptions (Rosen's setting \cite{rosen1965existence}):
\begin{enumerate}[(i)] \label{assm}
\item set $\mathcal{X} \subset \mathbb{R}^n$ is convex and compact.
\item Function $g_i: \mathbb{R}^n \rightarrow \mathbb{R}, i=1,...,M$ and $s: \mathbb{R}^n \rightarrow \mathbb{R}^{m_0}$ are convex and continuously differentiable.
\item The cost function of every player $J_i(x^i,x^{-i})$ is continuously differentiable in $x \in \mathcal{X}$.
\item The cost function of every player $J_i(x^i,x^{-i})$ is pseudo-convex in $x^i$ for every given $x^{-i}$.
\end{enumerate}

Define the Lagrangian function for each player as:
\begin{align}
    \mathcal{L}_i := J_i(x^i&,x^{-i})+ \mu_i^Th_i(x^i) + \nonumber\\
     &+\lambda_i^T \cdot g_i(x^i) + \sigma_i^T s(x^i,x^{-i})
\end{align}

where, $\mu_i \in \mathbb{R}^{k_i}$ are the Lagrange multipliers of the equality constraints of player $i$, $\lambda_i \in \mathbb{R}_+^{m_i}$ and $\sigma_i \in \mathbb{R}_+^{m_0}$ are the Lagrange multipliers of the private and shared inequality constraints of player $i$ respectively. For a point $x \in \mathcal{X}$ to be a GNE, the following KKT conditions have to be satisfied:
\begin{equation} \label{KKT}
\begin{split}
    & \nabla_{x^i}\mathcal{L}_i =0, \quad \forall i =1,...,M \\
    & 0 \leq \lambda_i \perp g_i \leq 0, \quad \forall i =1,...,M \\
    & 0 \leq \sigma_i \perp s \leq 0, \quad \forall i =1,...,M \\
    & \lambda_i \geq 0,\sigma_i \geq 0 \quad \forall i =1,...,M \\
    & h_i = 0, \quad \forall i =1,...,M \\
\end{split}
\end{equation}
To determine the Generalized Nash Equilibrium (GNE) in a Generalized Nash Equilibrium Problem (GNEP), we rely on certain conditions. Specifically, assuming that a suitable Constraint Qualification (CQ) holds—such as Slater’s condition or linear independence CQ—the authors in \cite{dreves2011solution, bueno2019optimality, facchinei2010generalized}  proved that :  

\begin{itemize}
    \item If a set of variables \((\bar{x}, \{\bar{\mu}_i\}_{i=1}^N, \{\bar{\lambda}_i\}_{i=1}^N,\{\bar{\sigma}_i\}_{i=1}^N)\) satisfies the Karush-Kuhn-Tucker (KKT) conditions, then \(\bar{x}\) is a solution to the GNEP.
    \item Conversely, if \(\bar{x}\) is indeed a solution to the GNEP, then there exists a corresponding set of Lagrange multipliers \((\{\bar{\mu}_i\}_{i=1}^N, \{\bar{\lambda}_i\}_{i=1}^N,\{\bar{\sigma}_i\}_{i=1}^N)\) that satisfy the KKT conditions when evaluated at \(x = \bar{x}\).
\end{itemize}

In other words, solving for the GNE can be done by checking the KKT conditions, and any valid GNE must have an associated set of multipliers that satisfies these conditions.


\section{Solutions To The GNEP} \label{sol}
Rosen introduced the concept of the \emph{normalized} solution for the Generalized Nash Equilibrium Problem (GNEP) in \cite{rosen1965existence}. Under the previously stated assumptions, the normalized solution corresponds to a scenario where the Lagrange multipliers associated with the shared constraints are \emph{identical} for all players:
\begin{equation} \label{shared_sigma}
    \sigma^1 = \dots = \sigma^N = \sigma.
\end{equation}
Rosen showed that a unique normalized solution exists under the specified conditions. Subsequent works, including \cite{harker1991generalized,facchinei2007generalized}, showed that reformulating the GNEP as a Quasi-Variational Inequality (QVI) problem inherently reduces the solution set to the normalized solution~\cite{facchinei2007generalized}. While having many numerical advantages, and of-the-shelves numerical solver that can be used, loosing is a byproduct that is unavoidable in his approach.

As a result, the normalized solution has become the de facto standard in solving GNEPs, \emph{to the extent that many works do not explicitly mention this assumption}. A popular method to calculate the normalized solution is by reformulating the GNEP to a Mixed Complementarity Problem (MCP) while assuming the normalized solution and using a numerical solver, such as PATH \cite{dirkse1995path}.

\subsection{The issue with normalized solutions: A simple example}

\begin{figure}[ht!] 
\centering 
\includegraphics[scale=0.4]{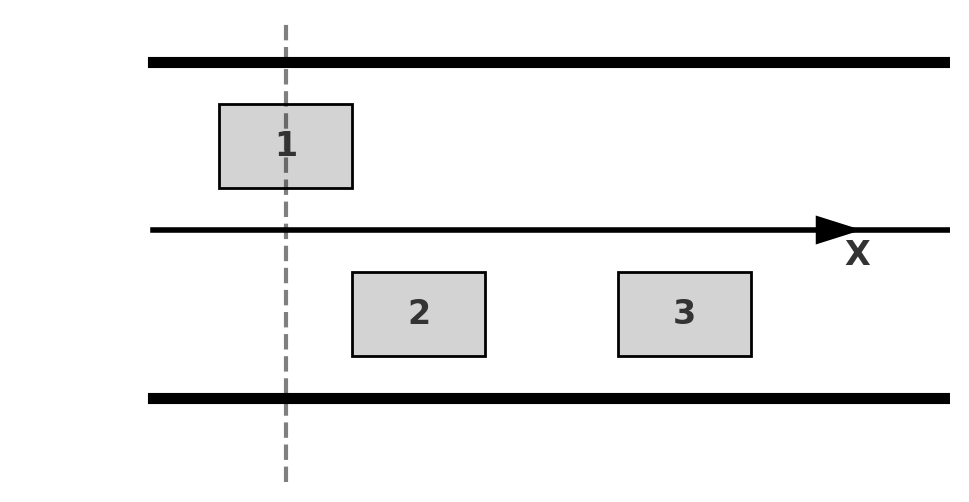} 
\caption{Illustration of the 1D racing example.} 
\label{fig:race} 
\end{figure}

A normalized solution might is a constraint on the solution that may not always produce the desired strategy, as we show in the following example. Consider a single-step, finite-horizon, discrete-time, general-sum, open-loop dynamic game in a racing scenario involving three cars on a two-lane track. Each car is a player in the generalized Nash equilibrium problem (GNEP), where it controls its velocity along a one-dimensional lane. Car 1 occupies the first lane, while Cars 2 and 3 share the second lane (Figure~\ref{fig:race}). $x_i$ and $v_i$ represent the final position and velocity of the car $i$, respectively. The game for player $i \in \{1,2,3\}$:

\begin{equation} \label{exm:setup}
\begin{alignedat}{3}
    &\min_{x_i, v_i} J_i(x_i,v_i,x^{-i}) \\
    &\text{s.t.} \\
    &x_i = x_i(0) + v_i\cdot \Delta t  \\
    &x_2 \leq x_3 \\
\end{alignedat}
\end{equation}

where $\Delta t = 1[sec]$, and $(x_1(0),x_2(0),x_3(0))=(0,0.5,0.75)$. The cost functions of the players are:
\begin{equation} \label{exm:cost}
\begin{alignedat}{3}
    &J_1 = -x_1 + x_2 + \frac{1}{2}v_1^2 \\
    &J_2 = -x_2 + x_1 + \frac{1}{2}v_2^2 \\
    &J_3 = -x_1 + x_2 + \frac{1}{2}v_3^2 
\end{alignedat}
\end{equation}
The objective of Car 1 is to advance farther than Car 2 while minimizing control effort. Similarly, Car 2 aims to advance farther than Car 1 while minimizing its control effort. Car 3, however, seeks to assist Car 1 in winning the race while keeping its control effort minimal. A collision avoidance constraint between Cars 2 and 3 makes it possible for car 3 to affect car 2 trajectory. 

Figure \ref{fig:race} illustrates the initial conditions of the race. Solving the above dynamic game for the normalized GNE gives the following results (See \cite{pustilnik2025generalizednashequilibriumsolutions} for a full derivation):
\begin{equation} \label{norm_exp:sol2}
\begin{split}
    &x_1 = x_1(0) + \Delta t^2 = 1.0 \\
    & x_2=x_3=\frac{x_2(0) + x_3(0)+\Delta t^2}{2} = 1.125
\end{split}
\end{equation}
The solution obtained from the normalized approach is counterintuitive. In the normalized solution, Car 3 does not block Car 2 as expected. Instead, it moves forward, worsening its own cost function. If Car 3 remained stationary, it would achieve a lower cost.

The normalized GNE, constrained by the shared $\sigma$s between all players, gives a solution that car 3 would not have chosen under any other reasonable condition. It can be easily verified that any solution of the form $x_2=x_3 \in [0.5,1.5]$ is a GNE. Therefore, the normalized solution is just \emph{one possible option} for Car 3, and it is not the most appealing.
~\\

\subsection{Non-Normalized Solutions}

A non-normalized solution of a GNEP is a solution $x \in \mathcal{X}$ such that the KKT conditions hold and the Lagrange multipliers of the shared constraints are \emph{not} the same for all players.

In general, DG may have many possible GNE solutions and selecting the GNE strategy that will give you the best final result should be done by a suitable consideration.

We proposes a novel method to exploit the MCP reformulation of a GNEP to calculate non-normalized solutions. Recall the KKT conditions from the original GNEP formulation (\ref{KKT}). The main idea was to compute a single set of fictitious shared Lagrange multipliers and introduce a strictly positive scaling factor for each player's shared constraints. Now let's define for every player $i$ a diagonal matrix of weights:
\begin{equation}
\begin{aligned}
\label{factors}
    A_i &\in \mathcal{D}^+_m, \\
    \mathcal{D}^+_m &:= \left\{ D \in \mathbb{R}^{m \times m} \mid 
    D = \text{diag}(d_1, \dots, d_m), \right. \\
    &\quad \left. d_i > 0, \forall i = 1, \dots, m \right\}.
\end{aligned}
\end{equation}
Each $A_i$ is used to scale the fictitious shared Lagrange multipliers for player $i$ to produce the actual Lagrange multipliers. Incorporating these factor matrices into the Lagrangian of each player gives:
\begin{align}
    \mathcal{L}_i := J_i(x^i&,x^{-i})+ \mu_i^Th_i(x^i) + \nonumber\\
     &\lambda_i^T \cdot g_i(x^i) + (A_i \sigma)^T s(x^i,x^{-i})
\end{align}
where $\sigma \in \mathbb{R}^{m_0}$ is the fictitious Lagrangian multipliers vector which is shared by all players.

The KKT conditions in \ref{KKT} are reformulated as:
\begin{equation} \label{modifiedKKT}
\begin{aligned}
    & \nabla_{x^i}\mathcal{L}_i(x^i, x^{-i})= 0, \quad \forall i \in \{1, \dots, M\}, \\
    & 0 \leq \lambda_i \perp g_i \leq 0, \quad \forall i \in \{1, \dots, M\}, \\
    & 0 \leq \sigma \perp s \leq 0, \\
    & \lambda_i \geq 0,h_i=0, \quad \forall i \in \{1, \dots, M\}, \\
    & \sigma \geq 0
\end{aligned}
\end{equation}

In this formulation, a single vector of fictitious Lagrange multipliers $\sigma$ is computed, with the scaling applied differently for each player using the factor matrices $A_i$ producing the actual Lagrange multipliers for each player. 

 In \cite{pustilnik2025generalizednashequilibriumsolutions} we prove that 
for a given a set of matrices $\{A_i\}_{i=1}^M$ as defined in (\ref{factors}) if one is able to compute the tuple $(\bar{x}, \{\bar{\mu}_i\}_{i=1}^N, \{\bar{\lambda}_i\}_{i=1}^N, \bar{\sigma})$ that solves the KKT conditions in (\ref{modifiedKKT}), then $\bar{x}$ is a solution to the GNEP.
This result provides a method for calculating non-normalized solutions to a GNEP using the tools developed for normalized solutions, and specifically numerical solvers of MCP. The solution process uses the standard procedure of solving GENP \cite{liu2023learning} using the MCP formulation 
 by modifying the players Lagrangian as presented above. 
 
 If all factor matrices $A_i$ are identical, the solution reduces to the normalized GNE. The factor matrices $A_i$ can be interpreted as representing the relative ``aggressiveness" of each player: lower values in $A_i$ for active shared constraints indicate a reduction in the corresponding Lagrange multipliers, allowing the player to improve their cost relative to others.
 
 Furthermore, and without loss of generality, the factor matrix for player 1 can be defined as $A_1=I_{m_0}$. This is due to the fact that the factor matrices are all relative. The same solution $(\bar{x}, \{\bar{\mu}_i\}_{i=1}^N, \{\bar{\lambda}_i\}_{i=1}^N, \{\bar{\sigma}_i\}_{i=1}^N)$ is valid, if all factor matrices are divided by some factor and $\bar{\sigma}$ is multiplied by the same factor - this will ensure that the Lagrange multipliers $\bar{\sigma_i}$ will stay the same.

 As mentioned, the most popular tools to solve a GNEP are QVI or MCP reformulations, which produce the normalized solution when used as discussed in the previous sections.
While these formulations offer many mathematical and numerical advantages, including compatibility with widely available solvers, generalizing these formulations to allow for non-normalized solutions can be beneficial in many cases.

\section{Results: A Two-Cars Racing Problem} \label{num_2cars}
One of the most popular approaches to solving an autonomous car racing problem is by using a Dynamic Game (DG) approach. In a 2 cars race, each car has a unique dynamic model and a cost function to minimize. A DG formulation is set up for each car with some shared prediction horizon. The game’s shared constraint is Collision Avoidance (CA) between both players. The DG is solved by finding the GNE of the problem, usually using a MCP formulation solved with a numerical solver, and the shared constraints Lagrange multipliers are assumed to be the same \cite{zhu2023sequential,liu2023learning}. This problem often has multiple GNEs, and there is no clear reason to prefer the normalized solution over others.

A kinematic bicycle model for each car is used, with the following states and inputs:
\begin{equation} \label{states}
\begin{aligned}
    & x^i_k = [v, \psi, s, t, X, Y], \quad i=1,2 \quad k = 0,...,N \\
    & u^i_k = [u_a, u_\delta], \quad i=1,2 \quad k = 0,...,N-1
\end{aligned}
\end{equation}
where, $v$ represents the total velocity, $\psi$ is the angular state relative to the center-line, and $s, v_t$ are the arc-length position of the car in the local Frenet reference frame relative to the center line of the track and lateral deviation from the center-line, respectively. $X$ and $Y$ are the inertial position of the car. $u_a$ is the longitudinal acceleration and $u_\delta$ is the steering of the front wheels $N=10$ is the prediction horizon. Finally, the time step is set to $\Delta t=0.1[sec]$. 

To illustrate the effectiveness of the proposed method, the same dynamic model and cost function is used for both players. The Dynamic Game formulation for player $i \in {1,2}$ is given by:
\begin{equation}
\begin{aligned}
    &\min_{x^i, u^i} -s^i_N + s^{-i}_N + \frac{\beta}{2} \sum_{k=0}^{N-1} \|u^i_k\|^2 \\  
    &\text{s.t.} \quad x^i_{k+1} = f(x^i_k, u^i_k), \quad x^i_k \in \mathcal{X}, \quad u^i_k \in \mathcal{U}, \\  
    & (X^1_k - X^2_k)^2 + (Y^1_k - Y^2_k)^2 \geq d_{safe}, \quad k = 1, \dots, N-1,
\end{aligned}
\nonumber
\end{equation}
where, $f$ is the dynamics function, $\mathcal{X}$ is the feasible set of the states, $\mathcal{U}$ is the feasible set of the inputs, and $d_{safe} = 0.4[m]$ is the minimal distance between both vehicles, and $\beta = 10^{-1}$. As can be seen, both cars have the same characteristics and the same cost function properties. Each car aims to maximize its relative progress while minimizing control effort. 


To compute the GNE of this DG, an MCP formulation is numerically solved using the PATH solver \cite{dirkse1995path}. To define the Lagrangian of each car using the method presented in \ref{sol}, a different factor should be given to each collision avoidance (CA) constraint. In this problem, there are $N$ shared constraints for each car. To make the solution more tractable, the number of factors is lowered. The factor matrix of the first car is chosen to be $I_{N}$ as explained in section \ref{sol}, and the factor matrix of car 2 is reduced to $\alpha I_{N}$ which can further reduced to a scalar $\alpha$. The physical interpretation of $\alpha \in \mathbb{R}_{++}$ is the relative aggressiveness of player 2 relative to player 1. If $\alpha <1$ player 2 is more aggressive than player 1, and for $\alpha>1$ player 1 is more aggressive than player 2.

This means that there is a single parameter that factors the shared constraints Lagrange multipliers:
\begin{equation}
\begin{aligned}
\label{DG_formulation}
    \mathcal{L}_1& = -s^1_N + s^2_N + \frac{\beta}{2}\sum_{k=0}^{N-1}{\|u^1_k\|^2} + 
    \\&+\mu_1^T \cdot h(x^1, u^1) + \lambda_1^T\cdot g(x^1,u^1) + \\ 
    &+\sigma^T\cdot g_{CA}(x^1,x^2) \\
    \mathcal{L}_2& = -s^2_N + s^1_N + \frac{\beta}{2}\sum_{k=0}^{N-1}{\|u^2_k\|^2} + \\
    &+\mu_2^T \cdot h(x^2,u^2) + \lambda_2^T\cdot g(x^2,u^2) + \\ 
    &+ \alpha \cdot\sigma^T\cdot g_{CA}(x^1,x^2)
\end{aligned}
\end{equation}
where  $\mu_1, \mu_2$ are the Lagrange multipliers associated with the dynamics (equality) constraints  $h$. $\lambda_1, \lambda_2$ are the Lagrange multipliers associated with the state and input constraints $g$. $\sigma$ is the fictitious Lagrange multiplier vector associated with the CA constraint $g_{CA}$. The next step is to apply the stationary condition on each Lagrangian and complementary slackness conditions:
\begin{equation}
\begin{aligned}
&\frac{\partial \mathcal{L}_1}{\partial(x^1,u^1)} = 0,\quad \frac{\partial \mathcal{L}_2}{\partial(x^2,u^2)} = 0 \\
&x^1_{k+1} - f_1(x^1_k, u^1_k) = 0,\quad x^2_{k+1} - f_2(x^2_k, u^2_k) = 0 \\
& g_1(x^1,u^1) \leq 0,\quad  g_2(x^2,u^2) \leq 0 \\
& g_{CA}(x^1,x^2) \leq 0
\end{aligned}
\end{equation}
Solving the above KKT equations for any value of $\alpha \in (0,\infty)$ gives a valid GNE solution.

\subsection{Qualitative Results: Multi-Modal Game Solutions}

The solution set of two examples are presented here to illustrate the effectiveness and added value of our method. Each example corresponds to a specific initial state of the cars on different tracks.

\begin{figure}
    \centering
    \includegraphics[scale=0.29]{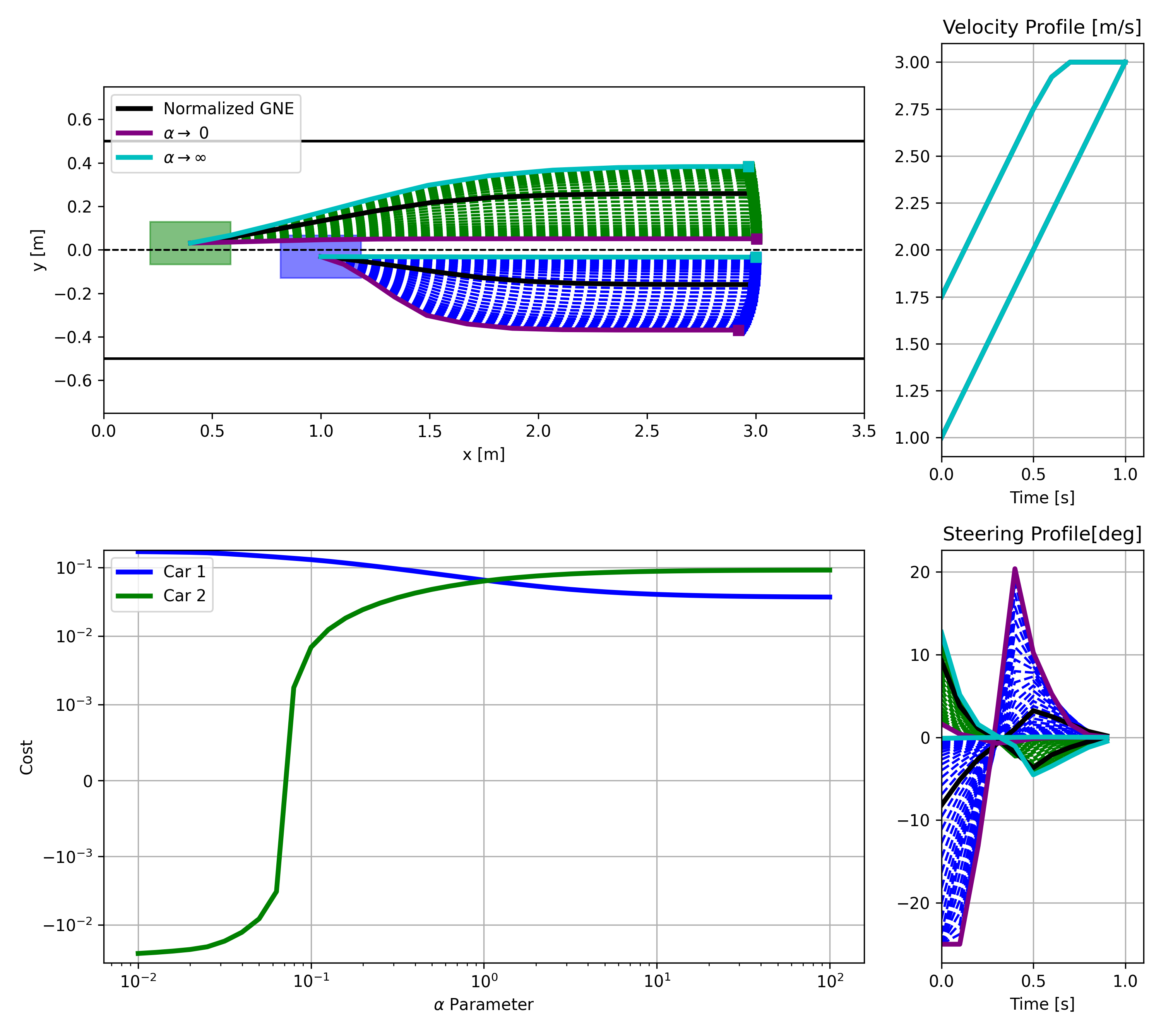}

    \caption{First Qualitative Example: Racing Problem Solutions on a Straight Track}
    \label{fig:race_straight}
\end{figure}

The first example is calculated on a straight track. Car 1 is ahead of car 2 but has a velocity disadvantage. Figure \ref{fig:race_straight} shows the solution of the race for different $\alpha$. The solution for $\alpha \rightarrow 0$ gives a straight trajectory (red line) for car 2 and this corresponds to car 2 being more aggressive than car 1. The solution for $\alpha \rightarrow \infty$ gives a straight trajectory (cyan line) for car 1 and this corresponds to car 1 being more aggressive than car 2. The GNE solution changes continuously as $\alpha$ changes. The lower graph shows the cost function of both cars as function of $\alpha$. A car can improve its cost by playing more aggressively. On the left side of the graph the velocity and steering profile of the cars are presented. The black line represent the normalized solution.

\begin{figure}
    \centering
    \includegraphics[scale=0.29]{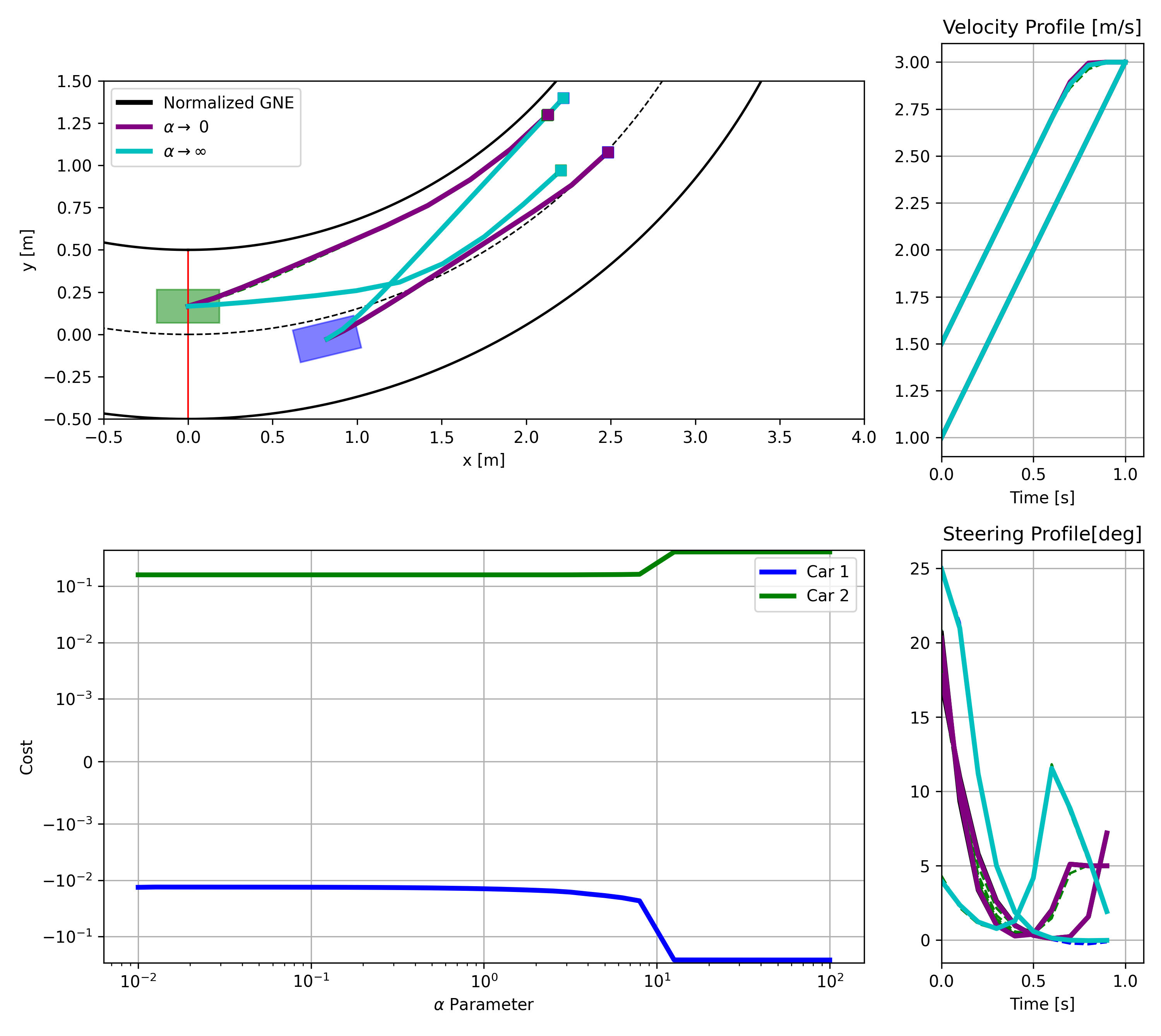}

    \caption{Second Qualitative Example: Racing Problem Solutions on a Curved Track}
    \label{fig:race_circle}
\end{figure}

The second example is calculated on a curved track. Car 1 is ahead of car 2 but has a velocity disadvantage. Figure \ref{fig:race_circle} shows the solution of the race for different $\alpha$. In this example there are 2 types of possibles solutions. For $\alpha \rightarrow 0$ Car 2 is much more aggressive and cuts into the corner which enforces Car 1 to take the wider turn. On the other hand, for $\alpha \rightarrow \infty$ car 1 is much more aggressive and it cuts into the corner before car 2 can do it - this enforces car 2 to take the wider turn.

The GNE solution does not change continuously as $\alpha$ changes. The lower graph shows the cost function of both cars as function of $\alpha$. This multi modal solution was not possible if only the normalized solution is calculated. It can be seen that for $\alpha \approx10$ there is a jump in the costs which corresponds to the jump in the solution types. On the left side of the graph the velocity and steering profile of the cars are presented.

\subsection{Quantitative Results: A Numerical Study}

The examples above show the set of options a race car has at different situations of the race. Choosing the right strategy from all possible is essential to get an advantage. The more aggressive a car will act (meaning, choosing lower $\alpha$) the better its performance will be. To prove this, we present a numerical study of a racing simulation of 2 racing cars. Both cars race by solving a racing DG problem as presented in (\ref{DG_formulation}). 

The scenario investigated is on an L-shaped track where the cars race in the counter-clock direction. The initial state of the cars is randomly initialized along the track with velocities such that the trailing car (called ego) has a velocity advantage over the forward car (Opponent).

Two scenarios are investigated. In the first, both cars implement a strategy that is the result of the normalized solution ($\alpha = 1$). In the second scenario, the opponent continues to implement the normalized strategy (meaning, it solve the DG while assuming both car use a normalized solution) while the ego car solves the DG with non-normalized solution ($\alpha=0.05$).

In both cases, the goal of the ego car is to overtake the opponent during a $2[sec]$ race started from the randomized initial conditions. The simulation is \emph{closed-loop}, meaning that each player solves the DG at each time step, implements a single control input and solves the DG again to get the next time step.

\begin{figure}
    \centering
    \includegraphics[scale=0.14]{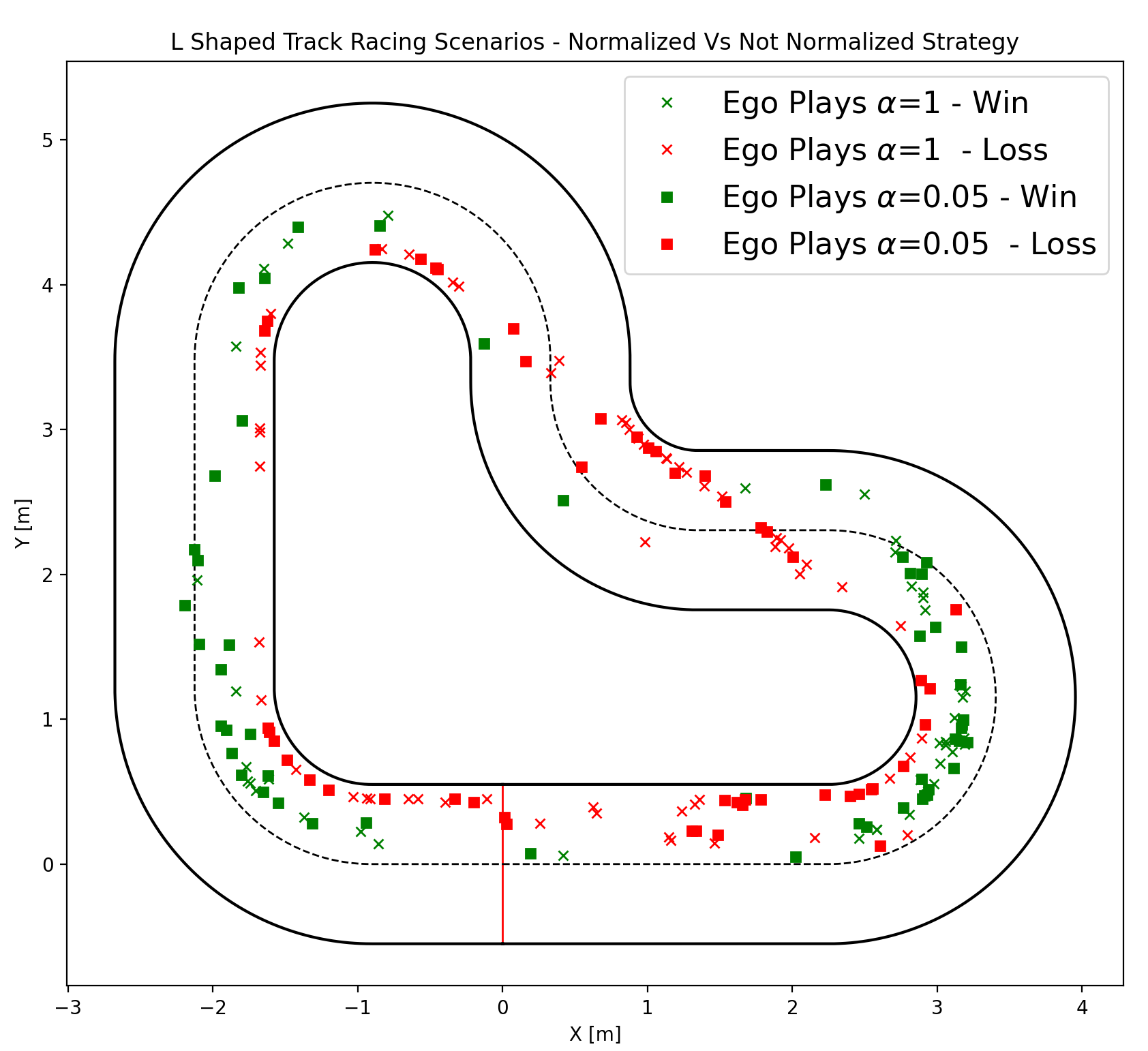}
    \caption{Results of 100 Monte Carlo simulations comparing two different strategies on an L-shaped track}
    \label{fig:MC_res}
\end{figure}

\begin{table}[]
    \centering
    \begin{tabular}{|c|c|c|}
        \hline
         & Ego Plays Normalized & Ego Plays Non-Normalized \\ 
        \hline
        Win Percentage & 43\% & 54\% \\ 
        \hline
    \end{tabular}
    \caption{MC simulation results summery comparing both strategies}
    \label{tab:results}
\end{table}

The ego has a top speed of $3[m/s]$ while the opponent has a top speed of $2.85[m/s]$, otherwise both cars have the same dynamic model and constraints. The initial state of the cars is randomized with properties presented in the following table:
\begin{table}[h]
    \centering
    \begin{tabular}{|c|c|}
        \hline
        \textbf{Description}  & \textbf{Range} \\ 
        \hline
        Opponent initial position  & $[0, L]$ \\ 
        \hline
        Ego initial position relative to opponent  & $[-1.75, -1.5]$ \\ 
        \hline
        Opponent initial velocity & $[1.0, 2.0]$ m/s \\ 
        \hline
        Ego initial velocity relative to opponent & $[0.25, 0.75]$ m/s \\ 
        \hline
        Ego lateral offset & $[-H/3, H/3]$ m \\ 
        \hline
        Opponent lateral offset (relative to ego) & $[- H/8,H/8]$ m \\ 
        \hline
    \end{tabular}
    \caption{Randomized Parameters in the Monte Carlo Simulation. $L$ is the track length and $H$ is the track half width.}
    \label{tab:random_params}
\end{table}

Table \ref{tab:results} summarizes the results of this study. When both cars use the normalized strategy (exhibiting similar levels of aggressiveness), the ego car overtakes the opponent in $43\%$ of the scenarios. However, when the ego car adopts a more aggressive strategy than the opponent, it wins $54\%$ of the scenarios—an $11\%$ \emph{increase} in win probability. Note that the advantage of the non-normalized strategy may be even greater, considering that certain randomized initial conditions may inherently prevent the ego car from winning, regardless of its strategy.
. 
Another visualization of these results is available in Figure \ref{fig:MC_res}. Each point on the track represents the initial condition of the ego car in a single racing scenario. A green scenario indicates a win for the ego car (successful overtaking of the opponent), while a red scenario indicates a loss (failure to overtake the opponent). This shows the effectiveness of the non-normalized strategy relative to the normalized one.

Overall, these results clearly demonstrate that a more aggressive strategy increases the likelihood of winning. While this conclusion is intuitive, proving it rigorously is nontrivial. The non-normalized solutions presented in this paper provide a systematic method for controlling the relative aggressiveness of players in a game and analyzing its impact on the final outcome.

Also, while the increase in the win percentage from 43\% to 54\% may not seem substantial, the theoretical upper bound on the optimal win rate in this setting is difficult to compute. We conjecture that, given the constraints of this scenario, the optimal win percentage when both players pursue non-normalized solutions is likely not significantly higher than 54\%.

Clearly, in real-time applications, an identification of the opponent's aggressiveness needs to be carried out and adjusted dynamically to optimize individual performance while avoiding crashes. Understanding the effects of time-varying opponent aggressiveness is an interesting direction for future research but is beyond the scope of this paper.

\section{Conclusion}
This paper introduces a novel method for extending existing approaches used to solve autonomous racing using the dynamic games (DG) approach. While traditional methods have been predominantly restricted to computing the normalized solution, the proposed framework allows for the computation of non-normalized solutions, significantly expanding the solution space in DGs while using the same powerful methods used to calculate the normalized solution. Furthermore, the examples demonstrate the effectiveness of the proposed framework in practical applications, highlighting its ability to identify diverse GNE solutions. Future work may include investigating criteria for choosing a GNE from the possible solutions.

In conclusion, this work expands the computational toolbox for DGs by enabling exploration beyond normalized solutions. These contributions pave the way for more robust and flexible applications of game-theoretic principles in complex, real-world scenarios.


\bibliographystyle{IEEEtran}
\bibliography{IEEEfull,bibliography.bib}

\begin{thebibliography}{10}
\providecommand{\url}[1]{#1}
\csname url@samestyle\endcsname
\providecommand{\newblock}{\relax}
\providecommand{\bibinfo}[2]{#2}
\providecommand{\BIBentrySTDinterwordspacing}{\spaceskip=0pt\relax}
\providecommand{\BIBentryALTinterwordstretchfactor}{4}
\providecommand{\BIBentryALTinterwordspacing}{\spaceskip=\fontdimen2\font plus
\BIBentryALTinterwordstretchfactor\fontdimen3\font minus \fontdimen4\font\relax}
\providecommand{\BIBforeignlanguage}[2]{{%
\expandafter\ifx\csname l@#1\endcsname\relax
\typeout{** WARNING: IEEEtran.bst: No hyphenation pattern has been}%
\typeout{** loaded for the language `#1'. Using the pattern for}%
\typeout{** the default language instead.}%
\else
\language=\csname l@#1\endcsname
\fi
#2}}
\providecommand{\BIBdecl}{\relax}
\BIBdecl

\bibitem{arrow2024existence}
K.~J. Arrow and G.~Debreu, ``Existence of an equilibrium for a competitive economy,'' in \emph{The Foundations of Price Theory Vol 5}.\hskip 1em plus 0.5em minus 0.4em\relax Routledge, 2024, pp. 289--316.

\bibitem{zhu2023sequential}
E.~L. Zhu and F.~Borrelli, ``A sequential quadratic programming approach to the solution of open-loop generalized nash equilibria,'' in \emph{2023 IEEE International Conference on Robotics and Automation (ICRA)}.\hskip 1em plus 0.5em minus 0.4em\relax IEEE, 2023, pp. 3211--3217.

\bibitem{spica2020real}
R.~Spica, E.~Cristofalo, Z.~Wang, E.~Montijano, and M.~Schwager, ``A real-time game theoretic planner for autonomous two-player drone racing,'' \emph{IEEE Transactions on Robotics}, vol.~36, no.~5, pp. 1389--1403, 2020.

\bibitem{dreves2011solution}
A.~Dreves, F.~Facchinei, C.~Kanzow, and S.~Sagratella, ``On the solution of the kkt conditions of generalized nash equilibrium problems,'' \emph{SIAM Journal on Optimization}, vol.~21, no.~3, pp. 1082--1108, 2011.

\bibitem{rosen1965existence}
J.~B. Rosen, ``Existence and uniqueness of equilibrium points for concave n-person games,'' \emph{Econometrica: Journal of the Econometric Society}, pp. 520--534, 1965.

\bibitem{harker1991generalized}
P.~T. Harker, ``Generalized nash games and quasi-variational inequalities,'' \emph{European journal of Operational research}, vol.~54, no.~1, pp. 81--94, 1991.

\bibitem{facchinei2007generalized}
F.~Facchinei, A.~Fischer, and V.~Piccialli, ``On generalized nash games and variational inequalities,'' \emph{Operations Research Letters}, vol.~35, no.~2, pp. 159--164, 2007.

\bibitem{liu2023learning}
X.~Liu, L.~Peters, and J.~Alonso-Mora, ``Learning to play trajectory games against opponents with unknown objectives,'' \emph{IEEE Robotics and Automation Letters}, vol.~8, no.~7, pp. 4139--4146, 2023.

\bibitem{bacsar1998dynamic}
T.~Ba{\c{s}}ar and G.~J. Olsder, \emph{Dynamic noncooperative game theory}.\hskip 1em plus 0.5em minus 0.4em\relax SIAM, 1998.

\bibitem{migot2019revisiting}
T.~Migot and M.-G. Cojocaru, ``Revisiting path-following to solve the generalized nash equilibrium problem,'' in \emph{International Conference on Applied Mathematics, Modeling and Computational Science}.\hskip 1em plus 0.5em minus 0.4em\relax Springer, 2019, pp. 93--101.

\bibitem{bueno2019optimality}
L.~F. Bueno, G.~Haeser, and F.~N. Rojas, ``Optimality conditions and constraint qualifications for generalized nash equilibrium problems and their practical implications,'' \emph{SIAM Journal on Optimization}, vol.~29, no.~1, pp. 31--54, 2019.

\bibitem{facchinei2010generalized}
F.~Facchinei and C.~Kanzow, ``Generalized nash equilibrium problems,'' \emph{Annals of Operations Research}, vol. 175, no.~1, pp. 177--211, 2010.

\bibitem{dirkse1995path}
S.~P. Dirkse and M.~C. Ferris, ``The path solver: a nommonotone stabilization scheme for mixed complementarity problems,'' \emph{Optimization methods and software}, vol.~5, no.~2, pp. 123--156, 1995.

\bibitem{pustilnik2025generalizednashequilibriumsolutions}
\BIBentryALTinterwordspacing
M.~Pustilnik and F.~Borrelli, ``Generalized nash equilibrium solutions in dynamic games with shared constraints,'' 2025. [Online]. Available: \url{https://arxiv.org/abs/2502.19569}
\BIBentrySTDinterwordspacing

\end{thebibliography}

\end{document}